\pdfoutput=1

\documentclass[11pt]{article}
\usepackage[most]{tcolorbox}
\usepackage{amsmath}
\usepackage{dsfont}
\usepackage[final]{acl}

\usepackage{times}
\usepackage{latexsym}
\usepackage{pifont}

\usepackage[T1]{fontenc}

\usepackage[utf8]{inputenc}

\usepackage{microtype}
\usepackage{hyperref}
\usepackage{inconsolata}

\usepackage{graphicx}
\usepackage{xspace}
\newcommand{\method}{\textsc{MathAgent}\xspace}

\usepackage{hyperref}
\usepackage{url}
\usepackage{hyperref}
\usepackage[most]{tcolorbox}
\usepackage{wrapfig}
\usepackage{graphicx,xcolor,float}
\usepackage{subcaption}
\usepackage{threeparttable}
\usepackage{algorithm}
\usepackage{pifont}
\usepackage[noend]{algorithmic}
\usepackage{colortbl}
\usepackage{color}
\usepackage{multirow}
\usepackage{tabularx}
\usepackage{float}
\usepackage{graphicx}
\usepackage{booktabs}
\usepackage{arydshln}
\usepackage{enumitem}
\usepackage{wrapfig}
\usepackage{caption}
\usepackage{graphicx}
\usepackage{makecell}
\usepackage{float} 
\usepackage{makecell}
\usepackage{tabularx}
\usepackage{twemojis}
\usepackage{amssymb,mathrsfs,amsmath}

\usepackage{pifont}
\usepackage[export]{adjustbox}
\usepackage{xcolor}
\usepackage{xspace}
\usepackage{shadowtext}
\usepackage{anyfontsize}
\usepackage{array}
\usepackage{caption}
\usepackage{subcaption}

\usepackage{xcolor}
\usepackage{graphicx}
\usepackage{pifont}
\usepackage{multirow} 
\usepackage{xcolor,colortbl}
\usepackage{times}
\usepackage{latexsym}
\usepackage{arydshln} 
\usepackage[acl,subfig]{definition}
\usepackage[detect-none]{siunitx}

\hypersetup{
    colorlinks=true,
    linkcolor=red,
    citecolor=cyan,
    filecolor=magenta,      
    urlcolor=magenta,
    }

\definecolor{bittersweet}{rgb}{1.0, 0.44, 0.37}
\definecolor{mygreen}{rgb}{0.29, 0.7, 0.48}
\definecolor{my_green}{RGB}{51,102,0}
\definecolor{my_yellow}{RGB}{255,165,0}
\definecolor{my_red}{RGB}{204, 0, 0}

\usepackage{pifont}
\definecolor{demphcolor}{RGB}{144,144,144}

\definecolor{mygray}{gray}{0.4}
\hypersetup{
    colorlinks=true,
    linkcolor=red,
    citecolor=cyan,
    filecolor=magenta,      
    urlcolor=magenta,
    }
\usepackage{xcolor} 
\usepackage{xcolor}

\usepackage[font=small,labelfont=bf]{caption}  
\usepackage{makecell}
\usepackage{tabulary}
\definecolor{ada_green}{rgb}{0,205,205}
\definecolor{glt_red}{rgb}{109,205,255}

\definecolor{backred}{RGB}{255, 190, 190}
\definecolor{backblue}{RGB}{210, 230, 250}
\definecolor{backgrey}{RGB}{220, 220, 220}

\definecolor{mygreen}{RGB}{184, 213, 118}
\definecolor{myred}{RGB}{215, 6, 84}
\definecolor{myblue}{RGB}{41, 115, 178}
\definecolor{shadecolor}{RGB}{237,237,237}

\title{\method: Leveraging a Mixture-of-Math-Agent Framework \\ for Real-World Multimodal Mathematical Error Detection}

\author{%
  Yibo Yan$^{1,2,3}$,
  Shen Wang$^{1}$,
  Jiahao Huo$^{2}$,
  Philip S. Yu$^{4}$,
  Xuming Hu$^{2,3,}$\footnotemark[2],
  Qingsong Wen$^{1,}$\footnotemark[2]\\
  \fontsize{9.0pt}{\baselineskip}\selectfont $^{1}$ Squirrel Ai Learning,  
  \fontsize{9.0pt}{\baselineskip}\selectfont $^{2}$ The Hong Kong University of Science and Technology (Guangzhou), \\
  \fontsize{9.0pt}{\baselineskip}\selectfont $^{3}$ The Hong Kong University of Science and Technology,
  \fontsize{9.0pt}{\baselineskip}\selectfont $^{4}$ University of Illinois at Chicago \\
   \texttt{\href{mailto:yanyibo70@gmail.com}{\{yanyibo70}, \href{mailto:qingsongedu@gmail.com}{qingsongedu\}@gmail.com}}, 
     \texttt{\href{mailto:xuminghu@hkust-gz.edu.cn}{xuminghu@hkust-gz.edu.cn}}
}

\begin{document}
\maketitle
\renewcommand{\thefootnote}{\fnsymbol{footnote}}
\footnotetext[2]{Corresponding authors.}
\renewcommand{\thefootnote}{\arabic{footnote}}
\begin{abstract} 
Mathematical error detection in educational settings presents a significant challenge for Multimodal Large Language Models (MLLMs), requiring a sophisticated understanding of both visual and textual mathematical content along with complex reasoning capabilities. Though effective in mathematical problem-solving, MLLMs often \textit{struggle with the nuanced task of identifying and categorizing student errors in multimodal mathematical contexts}. Therefore, we introduce \method, a novel Mixture-of-Math-Agent framework designed specifically to address these challenges. Our approach decomposes error detection into three phases, each handled by a specialized agent: an image-text consistency validator, a visual semantic interpreter, and an integrative error analyzer. This architecture enables more accurate processing of mathematical content by explicitly modeling relationships between multimodal problems and student solution steps. We evaluate \method on real-world educational data, demonstrating approximately 5\% higher accuracy in error step identification and 3\% improvement in error categorization compared to baseline models. Besides, \method has been successfully deployed in an educational platform that has served over one million K-12 students, achieving nearly 90\% student satisfaction while generating significant cost savings by reducing manual error detection.
\end{abstract}

\section{Introduction}
Multimodal Large Language Models (MLLMs) have revolutionized the landscape of artificial intelligence by enabling the integration and understanding of diverse data formats \cite{wu2023multimodal,xie2024large,yan2024urbanclip}. These models have demonstrated remarkable capabilities across various domains, from visual question answering to content generation and complex reasoning tasks \cite{yuan2025survey}. As education increasingly embraces digital transformation \cite{yan2025position,ye2025position}, the application of MLLMs to mathematical reasoning has emerged as a critical area of research, offering potential solutions to enhance teaching methodologies, provide personalized feedback, and support both educators and students in mathematical learning environments \cite{wang2024large,kuchemann2025opportunities,yan2024survey}.

\begin{figure}[!t]
    \centering
    \includegraphics[width=\linewidth,scale=1.00]{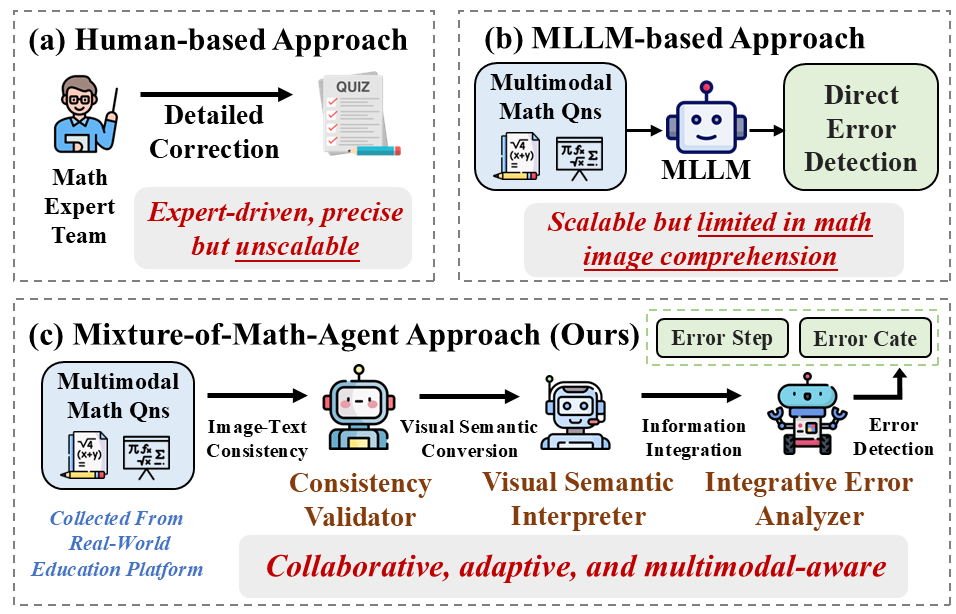}
    \caption{Comparison between previous human-based (a) and MLLM-based (b) paradigms vs. our proposed \method framework (c) for multimodal mathematical error detection.}
    \label{fig:paradigm_comparison}
    \vspace{-4mm}
\end{figure}

While significant progress has been made in utilizing MLLMs for mathematical problem-solving, a more practical and educationally valuable application lies in \textbf{mathematical error detection} \cite{yan2024errorradar,zheng2024processbench,song2025prmbench,yang2024supercorrect,li2024evaluating}. In real educational settings, identifying and categorizing students’ mathematical errors provides deeper insights into their conceptual understanding and learning gaps than merely evaluating final answers \cite{pepin2025mathematics,jiang2024llms}. Error detection is a significantly more challenging task for MLLMs compared to standard problem-solving, as it requires not only understanding the correct solution path but also analyzing the student’s flawed reasoning process. This task involves processing multiple inputs: the original problem (which may include multimodal elements), the correct solution, the student’s incorrect answer, and their detailed reasoning steps. The expected output comprises both error step identification (pinpointing exactly where the reasoning went wrong) and error categorization (classifying the type of misconception or mistake). This comprehensive analysis enables targeted educational interventions that address specific learning needs \cite{Chu2025LLMAF,yan2024survey}.

Existing error detection approaches face significant limitations when applied to real-world multimodal mathematical problems. \ding{182} As shown in Figure \ref{fig:paradigm_comparison}(a), traditional human-based approaches rely on expert teams to provide detailed corrections. While precise and pedagogically sound, these methods are inherently unscalable and cannot meet the growing demand for personalized feedback in digital learning environments \cite{li2024bringing}. \ding{183} As illustrated in Figure \ref{fig:paradigm_comparison}(b), MLLM-centric approaches, despite their computational scalability, exhibit suboptimal performance in mathematical image comprehension. For instance, symbolic representations in diagrams (\textit{e.g.,} misaligned coordinate systems) or mismatched text-image pairs (\textit{e.g.,} inconsistent geometric labels) often evade detection by MLLMs, leading to false predictions in error detection \cite{lu2023mathvista,zhang2024mathverse}.

To address these challenges, we propose and deploy \textbf{\method, a novel Mixture-of-Math-Agent framework specifically designed for multimodal mathematical error detection}. Drawing inspiration from expert-guided problem-solving practices \cite{chen2025symbolic,li2024smoa}, our framework decomposes the error detection workflow into three synergistic agents (refer to Figure \ref{fig:paradigm_comparison}(c)): an \textit{image-text consistency validator} to detect semantic consistency, a \textit{visual semantic interpreter} to extract structured expression from visual part of the problem, and an \textit{integrative error analyzer} that correlates all text-based inputs to pinpoint error locations and categorize misconception types. By explicitly modeling the interdependencies between textual problem formulations, visual mathematical objects, and solution steps, \method overcomes the aforementioned challenges inherent in both human-driven and MLLM-based approaches while maintaining computational tractability for real-world deployment.

Our contributions can be summarized as follows:

\ding{182} We introduce \method, the \textbf{first agent-based framework specifically designed for multimodal mathematical error detection}. Unlike previous paradigms that struggle with scalability, visual comprehension, and complex reasoning, \method leverages a novel mixture-of-agents approach, decomposing the task into multiple subtasks via specialized mathematical agents.

\ding{183} We validate our approach on\textbf{ data sampled from a real educational platform}, demonstrating performance improvements over baseline models. \method achieves approximately 5\% higher accuracy in error step identification and 3\% higher accuracy in error categorization, confirming its effectiveness in practical educational settings.

\ding{184} \method has been successfully \textbf{deployed in an educational platform that has served over one million K-12 students}. The system has achieved nearly 90\% student satisfaction rates while yielding estimated cost savings of approximately one million dollars by reducing the need for manual error detection, demonstrating both its practical utility and economic value.

\section{Related Work}
\vspace{-2mm}
\subsection{Mathematical Error Detection}
Mathematical error detection has evolved significantly from traditional rule-based systems to more sophisticated AI approaches \cite{li2024bringing,yan2024survey}. Early work focused on predefined error patterns and procedural mistakes in specific mathematical domains, such as arithmetic operations or algebraic manipulations \cite{rushton2018teaching}. With the advent of deep learning, researchers develop models capable of identifying more complex conceptual misunderstandings by analyzing student solution processes \cite{xu2024foundation}. Recent advances have leveraged LLMs to provide more nuanced error analysis and feedback generation, demonstrating promising results in understanding diverse student reasoning patterns \cite{li2025step,gao2024llm,li2024evaluating}. However, most existing research has primarily focused on text-based settings, with \textit{limited focus on multimodal contexts} where visual elements play a crucial role in problem representation \cite{yan2024errorradar}. \method extends the frontier of mathematical error detection by specifically addressing the challenges of multimodal mathematical reasoning, introducing a specialized agent-based framework.

\subsection{Agent for Mathematical Reasoning}
The application of agent-based approaches to mathematical reasoning has gained significant traction in recent years \cite{Chu2025LLMAF}. Initial efforts focused on single-agent systems that could execute predefined mathematical operations or follow structured solution procedures \cite{mitra2024orca,mei2024aios}. As LLMs advanced, researchers developed more sophisticated agents capable of step-by-step reasoning, self-verification, and even multi-step planning for complex mathematical problem-solving \cite{xiong2024building,wu2023mathchat,li2024ask}. Recent work has explored multi-agent frameworks where specialized agents collaborate on different aspects of mathematical reasoning, such as problem decomposition, solution planning, and verification \cite{gou2023tora,zhang2025correctness,xu2024ai}. However, existing agent-based systems for mathematical reasoning have \textit{primarily focused on problem-solving rather than error detection}, and few have adequately addressed the unique \textit{challenges posed by multimodal mathematical content}. Our \method represents a significant advancement in this domain by introducing a coordinated multi-agent system specifically designed for multimodal mathematical error detection.

See more related work in Appendix \ref{sec:more_related_work}.

\section{Our Proposed \method}
\subsection{Task Setting}
\label{sec:task_setting}

\begin{figure*}[thp]
    \centering
    \includegraphics[width=\textwidth]{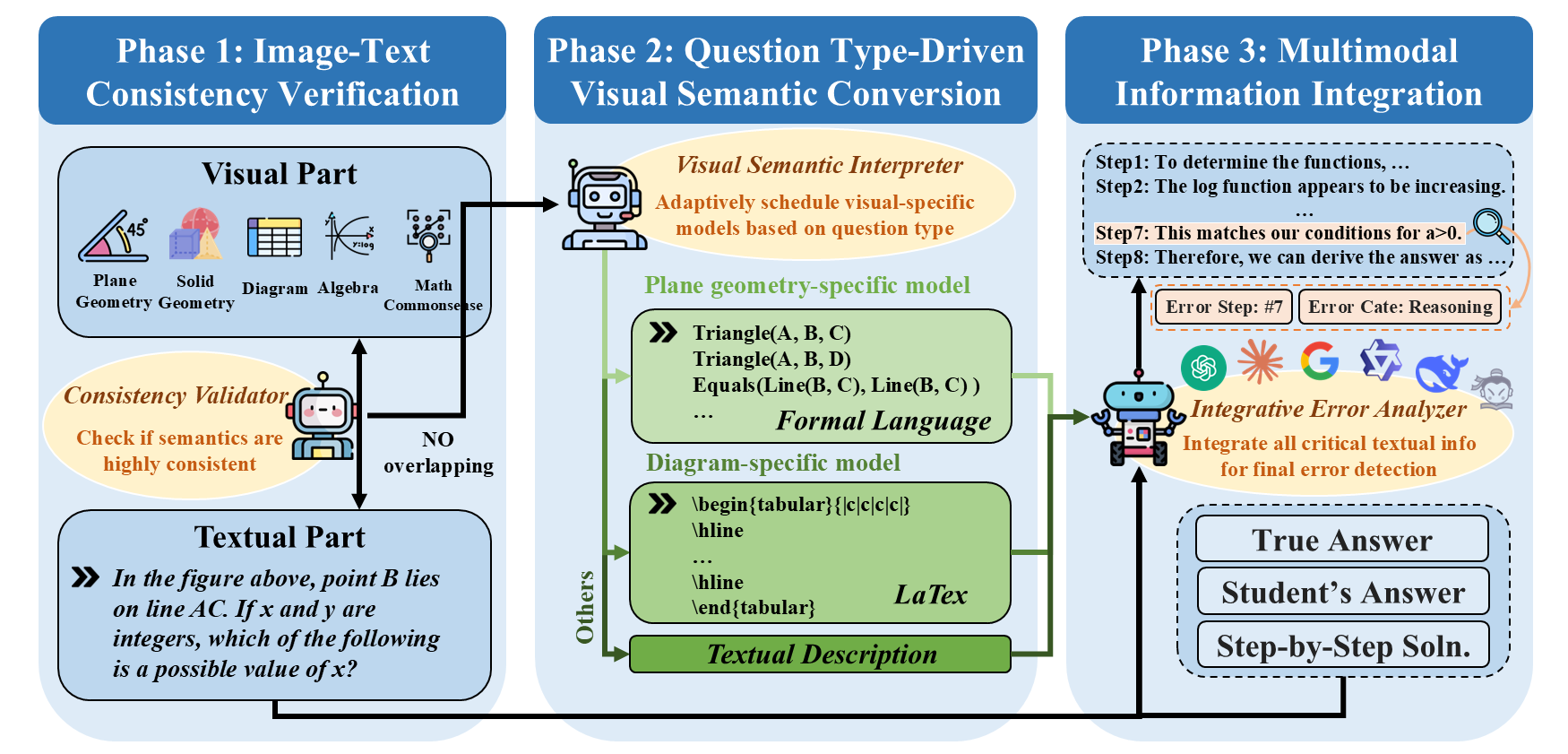}
    \caption{The framework of our proposed Mixture-of-Math-Agent for multimodal mathematical error detection.}
    \label{fig:pipeline}
    \vspace{-3mm}
\end{figure*}

We evaluate the framework's capability for multimodal error detection. The evaluation set contains $N$ samples.  For each sample $i$, input  $\mathcal{I}_i$ includes:

\begin{itemize}
    \item $Q_{\text{text}, i}$:  The textual problem statement.
    \item $Q_{\text{image}, i}$: The visual part of the problem.
    \item $A_{\text{correct}, i}$: The correct solution.
    \item $A_{\text{incorrect}, i}$:  An incorrect student solution.
    \item $\{S_{k, i}\}_{k=1}^{n_i}$: A sequence of  $n_i$ steps representing the student's step-by-step solution.
\end{itemize}

We define two subtasks as follows:

\textbf{Subtask 1: Error Step Identification.}  The goal is to identify the index,  $x_i$, of the \textit{first} incorrect step in the solution sequence  $\{S_{k, i}\}$.  Formally:
\[
x_i = \arg\min_{k} \{k \mid S_{k, i} \text{ is incorrect}\}
\]
\textbf{Subtask 2: Error Categorization.}  The goal is to classify the \textit{type} of error into one of five categories based on the first incorrect step:  VIS (Visual Perception), CAL (Calculation), REAS (Reasoning), KNOW (Knowledge), and MIS (Misinterpretation).  The error category is denoted as $C_{\text{error}, i}$. See details of error categories in Appendix \ref{sec:error_category_details}.

We use accuracy to evaluate performance.

\begin{itemize}
    \item \textbf{Error Step Identification Accuracy:}
    \vspace{-2mm}
    \[
    \text{Acc}_{\text{step}} = \frac{1}{N} \sum_{i=1}^{N} \mathbb{I}(x_i = G_{\text{step}, i})
    \]
    where $G_{\text{step}, i}$ is ground truth index of the first incorrect step, and $\mathbb{I}$ is the indicator function.

    \item \textbf{Error Categorization Accuracy:}
    \vspace{-2mm}
    \[
    \text{Acc}_{\text{cate}} = \frac{1}{N} \sum_{i=1}^{N} \mathbb{I}(C_{\text{error}, i} = G_{\text{error}, i})
    \]
    where $G_{\text{error}, i}$ is ground truth error category.
\end{itemize}

\subsection{Framework Overview}
\label{sec:framework_overview}

Our \method framework is designed for real-world multimodal mathematical error detection. As illustrated in Figure \ref{fig:pipeline}, the framework takes as input a multimodal mathematical problem (text and image), a correct answer, a student's incorrect answer, and their solution steps. The output is the identified error step and the corresponding error category. The framework operates in three sequential phases: Image-Text Consistency Verification (Sec.\ref{sec:phase1}), Question Type-Driven Visual Semantic Conversion (Sec.\ref{sec:phase2}), and Multimodal Information Integration (Sec.\ref{sec:phase3}). Each phase employs a specialized agent to perform a specific task.

\subsection{Phase 1: Image-Text Consistency Verification}
\label{sec:phase1}

\textbf{Motivation.}  Recent studies have demonstrated that MLLMs often exhibit lower performance in multimodal mathematical reasoning tasks when the image and text information are highly redundant \cite{zhang2024mathverse,lu2023mathvista}. This phenomenon highlights the current limitations of MLLMs in visual understanding and multimodal semantic alignment \cite{wu2024semantic,li2024multimodalalign}. Furthermore, in real-world educational settings, adaptively identifying high image-text consistency can improve efficiency, allowing us to bypass subsequent processing steps and directly proceed to error detection for highly overlapping problems.

\textbf{Methodology.} We introduce the \textit{Image-Text Consistency Validator}.  This agent takes the image and the textual description of the problem as input. It outputs a binary decision: whether the image and text are highly semantically consistent. The agent automatically determines the extent of semantic similarity between the image and text. Our system defaults to using GPT-4o\footnote{We used gpt-4o-2024-11-20.} as the agent for this phase. For example, if the image depicts a triangle with labeled angles and the text describes the same triangle and angles, the validator would output ``highly consistent.''

\subsection{Phase 2: Question Type-Driven Visual Semantic Conversion}
\label{sec:phase2}

\textbf{Motivation.} If the image and text information are not highly overlapping, we need an effective way to extract visual information for subsequent error detection. Inspired by recent advances in symbolic reasoning \cite{sullivan2024can,alotaibi2024graph,li2025proving}, we propose that MLLMs can adaptively dispatch specialized visual models based on the question type to convert visual information into a textual format. In particular, multimodal plane geometry problems, with their well-defined geometric relationships, are well-suited for conversion into formal language. Multimodal diagram problems, often involving tables or charts, are best represented using \LaTeX. Other types are converted into textual descriptions.

\textbf{Methodology.} We propose the \textit{Visual Semantic Interpreter}. This agent takes the image and the question type as input, and its output is a text-based representation of the visual information, tailored to the specific question type. The agent first determines the question type (\textit{e.g.,} plane geometry, diagram, algebra) and then selects the appropriate conversion method. Our system defaults to using corresponding visual-specific models\footnote{Refer to Appendix \ref{sec:visual_specific_models} for details.} as the agent for this phase. For instance, if the image is identified as a plane geometry setting, the interpreter might output a formal language representation like ``Triangle(A, B, C), Angle(BAC, 45), Line(AB, 5).''

\subsection{Phase 3: Multimodal Information Integration}
\label{sec:phase3}

\textbf{Motivation.}  Based on the extracted visual information from the previous phase, a comprehensive integration of all available information is crucial for accurate error localization. This phase must combine the problem's content, the student's incorrect answer, and their reasoning steps to pinpoint the cause of the error. The agent in this phase is directly responsible for the output of the two subtasks: error step identification and error categorization. Our system is designed to be compatible with any MLLM for inference, leveraging the increasingly powerful information integration capabilities of modern LLMs \cite{an2024make}.

\textbf{Methodology.} We introduce the \textit{Integrative Error Analyzer}. This agent takes  as input the problem's textual description, the converted visual information, the true answer, the student's answer, and the student's step-by-step solution. It outputs the identified error step and the error category. The agent first integrates all textual information and then analyzes the student's solution step-by-step, comparing it against the correct solution path. The agent for this phase is a flexibly selectable MLLM. For example, given a student's incorrect calculation in a geometry problem, the analyzer might output ``Error Step: \#3'' and ``Error Category: Calculation''.

\begin{table*}[htbp]
    \centering
    \caption{Main result of baseline MLLMs and corresponding \method framework. We denote STEP and CATE for error step identification and error categorization, the two subtasks of error detection, respectively, in Section \ref{sec:result}.}
    \label{tab:main_result}
    \small
    \renewcommand\tabcolsep{2.4pt}
    \begin{tabular}{lllllllll}
    \toprule
    \multirow{2}[2]{*}{\textbf{Model}} & \multirow{2}[1]{*}{\textbf{Error Step}} & \multicolumn{6}{c}{\textbf{Error Categorization}} & \multirow{2}[2]{*}{\textbf{Average}} \\ 
    \cmidrule(r){3-8}
    & \textbf{Identification} & \textbf{VIS} & \textbf{CAL} & \textbf{REAS} & \textbf{KNOW} & \textbf{MIS} &  \textbf{Overall} & \\ 
    
    \midrule
    \rowcolor{gray!15} GPT-4o \citep{openai2024gpt4o} & 55.10  & 46.30 & 50.40 & 64.90 & 9.20 & 46.30 & 53.08 & 54.09 \\

    w/ \textit{\method} & 59.50$^{\color{red}{4.4}\uparrow}$ & 48.40$^{\color{red}{2.1}\uparrow}$ & 55.00$^{\color{red}{4.6}\uparrow}$ & 63.90$^{\color{teal}{1.0}\downarrow}$ & 9.50$^{\color{red}{0.3}\uparrow}$  & 54.00$^{\color{red}{7.7}\uparrow}$  & 55.11$^{\color{red}{2.0}\uparrow}$ & 57.30$^{\color{red}{3.2}\uparrow}$ \\

    \rowcolor{gray!15} Gemini-Pro-1.5 \citep{reid2024gemini} & 52.00  & 9.10 & 46.80 & 62.70 & 31.90 & 13.00 & 44.51 & 48.26 \\

    w/ \textit{\method} & 57.90$^{\color{red}{5.9}\uparrow}$ & 15.70$^{\color{red}{6.6}\uparrow}$ & 48.50$^{\color{red}{1.7}\uparrow}$ & 61.30$^{\color{teal}{1.4}\downarrow}$ & 33.30$^{\color{red}{1.4}\uparrow}$  & 21.00$^{\color{red}{8.0}\uparrow}$  & 46.10$^{\color{red}{1.6}\uparrow}$ & 52.00$^{\color{red}{3.8}\uparrow}$ \\

    \rowcolor{gray!15} Claude-3.5-Sonnet \citep{claude35} &  50.20 & 35.70 & 48.40 & 64.80 & 21.00 & 11.40 & 49.50 & 49.85 \\

    w/ \textit{\method} & 55.10$^{\color{red}{4.9}\uparrow}$ & 40.10$^{\color{red}{4.4}\uparrow}$ & 55.30$^{\color{red}{6.9}\uparrow}$ & 62.70$^{\color{teal}{2.1}\downarrow}$ & 24.70$^{\color{red}{3.7}\uparrow}$  & 22.40$^{\color{red}{11.0}\uparrow}$  & 52.63$^{\color{red}{3.1}\uparrow}$ & 53.86$^{\color{red}{4.0}\uparrow}$ \\

    \rowcolor{gray!15} Qwen-VL-Max \citep{qwenmax} &  48.70 & 15.20 & 78.90 & 50.50 & 14.30 & 36.60 & 52.87 & 50.78 \\

    w/ \textit{\method} & 56.70$^{\color{red}{8.0}\uparrow}$ & 21.70$^{\color{red}{6.5}\uparrow}$ & 81.30$^{\color{red}{2.4}\uparrow}$  & 53.40$^{\color{red}{2.9}\uparrow}$ & 12.80$^{\color{teal}{1.5}\downarrow}$ &  36.60$^{\color{red}{0.0}\uparrow}$  & 55.80$^{\color{red}{2.9}\uparrow}$ & 56.25$^{\color{red}{5.5}\uparrow}$ \\

    \rowcolor{gray!15} InternVL2 \citep{chen2024internvl} &  54.40 & 33.40 & 92.40 & 25.10 & 10.90 & 8.10 & 49.46 & 51.93 \\

    w/ \textit{\method} & 56.30$^{\color{red}{1.9}\uparrow}$ & 38.80$^{\color{red}{5.4}\uparrow}$ & 85.30$^{\color{teal}{7.1}\downarrow}$ & 36.80$^{\color{red}{11.7}\uparrow}$  & 19.00$^{\color{red}{8.1}\uparrow}$  & 13.70$^{\color{red}{5.6}\uparrow}$  & 52.83$^{\color{red}{3.4}\uparrow}$ & 54.57$^{\color{red}{2.6}\uparrow}$ \\

    \rowcolor{gray!15} LLaVA-NEXT \citep{liu2024llavanext} &  48.44 & 7.10 & 86.00 & 32.00 & 7.60 & 0.80 & 45.08 & 48.44 \\

    w/ \textit{\method} & 57.60$^{\color{red}{5.8}\uparrow}$ & 15.70$^{\color{red}{8.6}\uparrow}$ & 84.50$^{\color{teal}{1.5}\downarrow}$ & 45.10$^{\color{red}{13.1}\uparrow}$  & 8.30$^{\color{red}{0.7}\uparrow}$  & 3.80$^{\color{red}{3.0}\uparrow}$  & 51.05$^{\color{red}{6.0}\uparrow}$ & 54.32$^{\color{red}{5.9}\uparrow}$ \\

    \midrule
    Average Improvement & $\color{red}{5.2}\uparrow$  & $\color{red}{5.6}\uparrow$ & $\color{red}{1.2}\uparrow$ & $\color{red}{3.9}\uparrow$ & $\color{red}{2.1}\uparrow$ & $\color{red}{5.9}\uparrow$ & $\color{red}{3.2}\uparrow$ & $\color{red}{4.2}\uparrow$ \\

    \midrule
    \rowcolor{blue!15} Human &  81.60 & 70.30 & 86.00 & 63.50 & 53.40 & 62.00 & 72.23 & 76.91 \\

    \bottomrule
    \end{tabular}
    \vspace{-4mm}
\end{table*}

\section{Experiment}
\vspace{-2mm}
\subsection{Experiment Settings}

\textbf{Dataset.} The dataset consists of a carefully curated collection of 2,500 multimodal mathematical questions sourced from real student problem-solving data on educational platforms. Each entry in this evaluation dataset has been meticulously selected by educational experts to ensure high quality, free from issues such as erroneous question design. The student responses represent the most frequent incorrect answers corresponding to each question. Furthermore, the erroneous steps and error category labels for each question have been determined through discussions among at least three experienced educational specialists. The dataset predominantly features plane geometry problems, supplemented by solid geometry, diagrams, algebra, and mathematical commonsense questions. Refer to Appendix \ref{sec:dataset_details} for more dataset details.

\textbf{Models.} We select representative MLLMs (See sources in Appendix \ref{sec:model_sources}) that have demonstrated effectiveness in recent studies \cite{yan2024errorradar,wang2024measuring,zhang2024mathverse}: InternVL-2 76B \citep{chen2024internvl}, LLaVA-NEXT 72B \citep{liu2024llavanext}, Qwen-VL-Max \citep{qwenmax}, Claude-3.5-Sonnet \citep{claude35}, Gemini-Pro-1.5 \citep{reid2024gemini}, and GPT-4o \citep{openai2024gpt4o}. These MLLMs are already deployed on the educational platform, allowing for a direct comparison of the gains achieved by \method. In our experiments, directly applying each MLLM to error detection serves as a baseline. We then evaluate the effectiveness of the \method framework by systematically decomposing the complex reasoning task, with the agent in Phase 3 retaining the baseline MLLM. Additionally, we engage evaluators with a background in education to conduct corresponding human evaluations, aiming to assess the gap between MLLM and human-level intelligence.

\subsection{Experimental Results \& Analysis}
\label{sec:result}

\textbf{Overall Performance Improvement with \method.} As shown in Table \ref{tab:main_result}, \method demonstrates significant performance improvements across both STEP and CATE subtasks. When integrated with various baseline MLLMs, \method consistently enhances their error detection capabilities, with an average improvement of 4.2\% across all models. Specifically, the framework boosts GPT-4o’s performance from 54.09\% to 57.30\% (3.2\% increase) and shows similar improvements for other models. This consistent enhancement across diverse architectures suggests that \method can address inherent challenges in multimodal mathematical error detection by systematically processing multimodal information.

\textbf{Differential Impact on STEP vs. CATE Tasks.} The \method framework yields more substantial improvements in STEP compared to CATE. Across all tested models, \method achieves an average improvement of 5.2\% in STEP tasks, while the enhancement for overall CATE tasks is 3.2\%. For instance, GPT-4o shows a 4.4\% improvement in STEP but only a 2.0\% improvement in CATE. This difference likely stems from \method’s information extraction and integration, which particularly benefits the error localization in sequential solution steps, while the more nuanced task of error categorization remains challenging.

\begin{figure}[!t]
    \centering
    \includegraphics[width=\linewidth,scale=1.00]{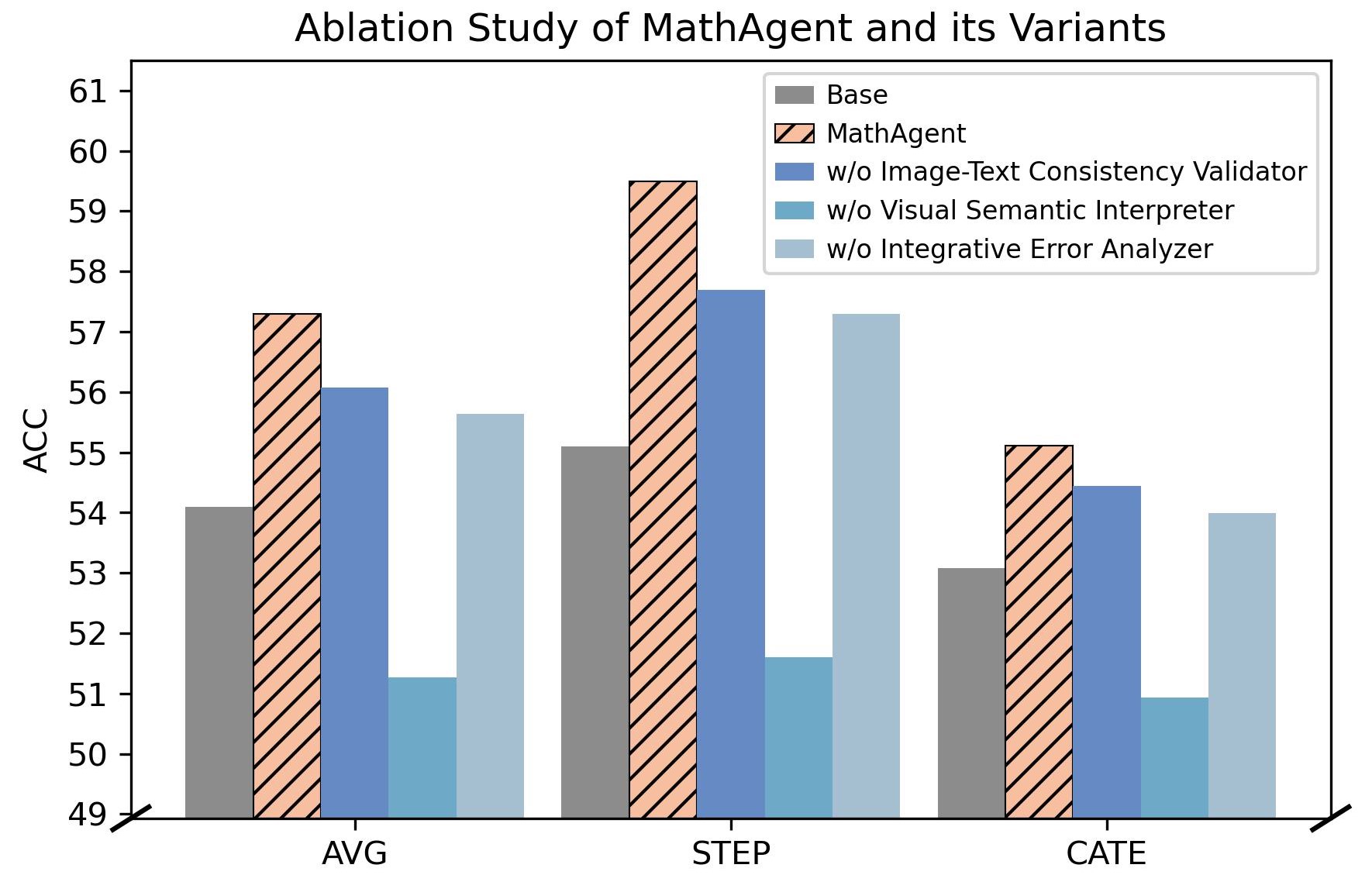}
    \caption{Ablation study of \method.}
    \label{fig:ablation}
    \vspace{-6mm}
\end{figure}

\textbf{Category-Specific Performance Variations.} \method demonstrates the most significant improvements in detecting VIS and MIS, with average enhancements of 5.6\% and 5.9\% respectively across all models. For example, Gemini-Pro-1.5 shows a remarkable 6.6\% improvement in VIS and 8.0\% in MIS categories when augmented with \method. In contrast, improvements in CAL, REAS, and KNOW are more modest at 1.2\%, 0.4\%, and 2.1\% respectively. This pattern highlights \method’s effectiveness in addressing multimodal integration challenges, as VIS and MIS errors fundamentally involve misalignments between visual information and problem interpretation.

\textbf{Gap Between \method and Human Performance.} 
Despite the notable improvements, \method still falls short of human-level performance in mathematical error detection. The best-performing \method-enhanced framework (GPT-4o at 57.30\%) remains significantly below human performance (76.91\%). The persistent performance gap underscores the inherent complexity of mathematical error detection, which requires sophisticated reasoning abilities, domain knowledge, and multimodal understanding.

\subsection{Ablation Study}

As depicted in Figure \ref{fig:ablation}, we evaluate performance of our \method framework and its ablative variants, using GPT-4o with the best overall performance as the base setting. We investigate three variants: (i) \textit{w/o Image-Text Consistency Validator}, which bypasses consistency check and processes all images in Phase 2; (ii) \textit{w/o Visual Semantic Interpreter}, which replaces question type-driven visual model scheduling with a unified captioning approach for all images; and (iii) \textit{w/o Integrative Error Analyzer}, which simply concatenates transcribed image information with student’s solution steps and answer, omitting the integration with the problem’s textual description. The results demonstrate that \method achieves the highest accuracy on both STEP and CATE tasks. Notably, the w/o Visual Semantic Interpreter variant exhibits the lowest performance, presumably because generic descriptions of abstract geometric images may omit crucial details like edge lengths and angle measures. Removing the Image-Text Consistency Validator also leads to a performance drop, suggesting that discrepancies between potentially flawed image transcriptions and textual problem description can introduce contradictory information, negatively impacting the complex reasoning process.



\section{Industrial Impact}
\vspace{-1mm}
\textbf{Error Detection Performance Enhancement in Real-World Educational System.} When deployed in educational platforms, \method has demonstrated remarkable improvements in error detection performance that directly translate to educational value. As a diagnostic tool, \method provides more precise feedback on student work, enabling targeted interventions. Furthermore, \method’s adaptive architecture optimizes computational resources by automatically filtering problems based on image-text consistency and selecting specialized visual models according to problem types.

\textbf{Student Satisfaction Rate Improvement.} A/B testing conducted on the educational platform reveals significant improvements in student satisfaction with \method-powered feedback systems. In a controlled study involving 10,000 K-12 students, \method-enhanced feedback received an over 90\% satisfaction rating, compared to 75\% for traditional MLLM-based feedback. These improvements in student experience demonstrate \method’s effectiveness as a pedagogically valuable tool that enhances the learning process.

We discuss more impact in Appendix \ref{sec:more_impact}.

\section{Conclusion}
\vspace{-2mm}
This paper presented \method, a novel and effective framework for multimodal mathematical error detection in real-world educational settings.  By leveraging a mixture-of-agent approach, \method overcomes the limitations of existing human-based and MLLM-centric methods, achieving superior performance in identifying and categorizing student errors. The successful deployment of \method on a large-scale educational platform, with improvements in accuracy, student satisfaction, and cost-effectiveness, underscores its significant technical and practical value.

\newpage
\clearpage
\section*{Limitations}
Despite the contributions demonstrated in our work, several limitations remain:
\begin{itemize}
\item [1.] The effectiveness of \method is contingent on the quality of the multimodal inputs. Poorly formatted or ambiguous problems may lead to inaccurate error detection. We will enhance our engineering pipeline to improve data cleaning and optimization processes, ensuring that input data is standardized and of high quality, which will lead to more accurate error detection.\par
\item [2.] While \method improves error detection accuracy, it may still struggle with a broader range of error categories beyond the five specified. We will collaborate with educational experts to develop a more comprehensive framework of error categories that aligns with student needs and encompasses a wider variety of mathematical errors. \par  
\item [3.] \method does not incorporate recent advancements in o1-like slow-thinking reasoning, which may enhance the depth of error analysis but could impact user feedback time in deployed systems. In the future, we will explore integrating user intent recognition to adaptively schedule fast and slow reasoning modes, providing students with comprehensive and timely error analysis based on their needs.\par
\end{itemize}

\section*{Acknowledgements}
This work was supported by NSF under grants III-2106758, and POSE-2346158; Guangdong Provincial Department of Education Project (Grant No.2024KQNCX028); Scientific Research Projects for the Higher-educational Institutions (Grant No.2024312096), Education Bureau of Guangzhou Municipality; Guangzhou-HKUST(GZ) Joint Funding Program (Grant No.2025A03J3957), Education Bureau of Guangzhou Municipality.
\bibliography{mathagent}

\clearpage
\appendix

\section{More Related Work}
\label{sec:more_related_work}

\subsection{Multimodal Large Language Model}
Current MLLMs adopt a similar framework, including a vision encoder, a connector, and an LLM backbone, which was initially proposed by LLaVA \cite{liu2024llava}. By training these components via visual instruction tuning, the vision embeddings extracted by the vision encoder are aligned with the word space of LLM through the connector \cite{raiaan2024review,shao2024survey}. Such a framework enables MLLMs to understand visual input such as images and video, while preserving the powerful reasoning and generation abilities of autoregressive LLMs \cite{deng2025following}. As a result, some MLLMs achieve state-of-the-art performance across a wide variety of multimodal tasks such as visual question answering \cite{manas2024improving,xiao2025videoqa}, image captioning \cite{bianco2023improving,patel2025alt}, video understanding \cite{zhou2024survey,huang2024vtimellm}, and more diverse tasks \cite{yan2024georeasoner,huo2024mmneuron}. On the other hand, with the development of o1-like systems in LLMs \cite{li2025system,zhong2024evaluation,jaech2024openaio1}, there is also a tendency to trigger the slow-thinking potentials of MLLMs \cite{yang2025r1,zhao2025r1,yao2024mulberry}. For example, Virgo \cite{du2025virgo} makes a preliminary exploration of multimodal slow-thinking systems by directly fine-tuning a capable MLLM with a small amount of textual long-form thought data, while Vision-o1 \cite{ni2024visual} proposes a multimodal multi-turn chain-of-thought framework to simulate human reasoning for MLLMs on ambiguous instructions. Furthermore, LlamaV-o1 \cite{thawakar2025llamav} uses a multiturn curriculum learning approach to facilitate MLLMs in incremental skill acquisition and problem-solving. Despite these efforts, the development of o1-like multimodal systems is still in its stages \cite{masterman2024landscape,chen2025towards,xu2025towards}, with significant problems such as overthinking \cite{cuadron2025danger,yang2025towards}, safety \cite{zhao2024survey,chen2025safeeraser,huo2025mmunlearner}, and hallucination \cite{sun2025hallucinations,zheng2024reefknot,zhou2024mitigating}.


\section{Error Category Details}
\label{sec:error_category_details}

The discrepancies within the five error categories are delineated as follows:
\begin{itemize}[leftmargin=*]
    \item[\ding{79}]  \textbf{Visual Perception Errors (VIS)}: These errors arise when there is a failure to accurately interpret the information contained within images or diagrams presented in the question due to visual issues.
    \item[\ding{79}]  \textbf{Calculation Error (CAL)}: These errors manifest during the calculation process, which may include arithmetic mistakes such as incorrect addition, subtraction, multiplication, or division, errors in unit conversion, or mistakes in the numerical signs between multiple steps.
    \item[\ding{79}]  \textbf{Reasoning Error (REAS)}: These errors occur during the problem-solving process when improper reasoning is applied, leading to incorrect application of logical relationships or conclusions.
   \item[\ding{79}]  \textbf{Knowledge Error (KNOW)}: These errors result from incomplete or incorrect understanding of the knowledge base, leading to mistakes when applying relevant knowledge points.
    \item[\ding{79}]  \textbf{Misinterpretation of the Question (MIS)}: These errors occur when there is a failure to correctly understand the requirements of the question or a misinterpretation of the question's intent, leading to responses that are irrelevant to the question's demands. For instance, if the question asks for a letter and a number is provided, or vice versa.
\end{itemize}

\section{Visual-Specific Models}
\label{sec:visual_specific_models}
In our deployed system, we employ specialized models tailored to different problem types to ensure optimal performance. For plane geometry problems, we utilize Inter-GPS\footnote{https://github.com/lupantech/InterGPS}, a groundbreaking geometry problem solver developed by \citet{lu2021inter}. As the first system capable of automatic program parsing and interpretable symbolic reasoning, Inter-GPS demonstrates its effectiveness through dual-channel processing: it employs rule-based text parsing for textual analysis and neural object detection for diagram interpretation, seamlessly converting problem texts and diagrams into formal language representations. Furthermore, its integration of theorem knowledge as conditional rules enables systematic, step-by-step symbolic reasoning.

When addressing diagram-based problems, particularly those involving tabular data, we implement StructTable-InternVL2-1B\footnote{https://github.com/Alpha-Innovator/StructEqTable-Deploy}, a sophisticated model developed by \citet{xia2024docgenome}. This end-to-end solution, known as StructEqTable, excels in visual table processing by accurately generating LaTeX descriptions from table images while simultaneously supporting multiple advanced functionalities, including structural extraction and question-answering capabilities, thereby significantly expanding its practical applications.

For general visual content processing beyond these specialized domains, we leverage the vit-gpt2-image-captioning model\footnote{https://huggingface.co/nlpconnect/vit-gpt2-image-captioning} to generate comprehensive and detailed image captions, ensuring robust performance across diverse visual understanding tasks.

\section{Dataset Details}
\label{sec:dataset_details}

\subsection{Dataset Statistics}
\label{sec:dataset_statistics}

Our evaluation dataset comprises 2,500 multimodal mathematical questions spanning diverse problem types and error categories. As illustrated in Figure \ref{fig:statistics}, the dataset is predominantly composed of Plane Geometry problems (62.4\%), followed by Algebra (11.5\%), Diagram problems (9.3\%), Math Commonsense (9.2\%), and Solid Geometry (7.6\%). This distribution reflects the prevalence of geometry-based problems in mathematical education that benefit significantly from visual representation and analysis.

The dataset captures a wide spectrum of error categories that students commonly encounter. Reasoning Errors constitute the largest proportion at 38.0\%, highlighting the challenges students face in logical deduction and proof construction. Calculation Errors account for 36.5\% of the dataset, representing arithmetic mistakes and computational inaccuracies. Visual Perception Errors make up 15.8\%, underscoring the importance of correctly interpreting visual elements in mathematical problem-solving. Knowledge Errors and Misinterpretation of Questions represent smaller but significant portions at 4.8\% and 4.9\% respectively.

The complexity of the problems is reflected in the reasoning steps required for solution, with an average of 7.6 steps per problem, ranging from a minimum of 3 to a maximum of 20 steps. The textual component of the problems varies considerably in length, averaging 168 characters, with the shortest problem containing just 13 characters and the most verbose extending to 719 characters. This variation in problem complexity and presentation provides a robust benchmark for evaluating \method’s performance across different mathematical contexts and difficulty levels.

\begin{figure}[!t]
    \centering
    \fontsize{8.2pt}{\baselineskip}\selectfont 
    \renewcommand\tabcolsep{1.0pt} 
    \renewcommand\arraystretch{0.8} 
    \begin{tabular}{lc} 
        \toprule
        \textbf{Statistic} & \textbf{Number} \\
        \midrule
        Total multimodal questions & 2,500 \\
        \midrule
        Problem Type &  \\
        ~- Plane Geometry & 1559 (62.4\%) \\
        ~- Solid Geometry & 191 (7.6\%) \\
        ~- Diagram & 233 (9.3\%) \\
        ~- Algebra & 288 (11.5\%) \\
        ~- Math Commonsense & 229 (9.2\%) \\
        \midrule
        Error Category &  \\
        ~- Visual Perception Error & 395 (15.8\%) \\
        ~- Calculation Error & 912 (36.5\%) \\
        ~- Reasoning Error & 951 (38.0\%) \\
        ~- Knowledge Error & 119 (4.8\%) \\
        ~- Misinterpretation of the Qns & 123 (4.9\%) \\
        \midrule
        Average Reasoning Step  & 7.6 \\
        Maximum Reasoning Step & 20 \\
        Minimum Reasoning Step & 3 \\
        Average Question Length  & 168 \\
        Maximum Question Length & 719 \\
        Minimum Question Length & 13 \\
        \bottomrule
    \end{tabular}
    \caption{Key statistics of dataset.}
    \label{fig:statistics}
\end{figure}

\subsection{Data Source}
The data used in this study originates from a real-world online education platform, ensuring its relevance and applicability to practical educational scenarios. This dataset is not synthetically generated; instead, it comprises authentic student submissions, including both correct and incorrect solutions. This provides a realistic representation of the types of errors students commonly make in a learning environment. Furthermore, the data includes a diverse range of mathematical problems, reflecting the breadth of topics covered in K-12 mathematics curricula. The use of real-world data enhances the ecological validity of our findings and ensures that the \method framework is evaluated on data that closely resembles the challenges encountered in actual educational settings. The platform anonymizes all student data to protect privacy, while preserving the integrity and richness of the information needed for effective error detection and analysis.

\section{Model Sources}
\label{sec:model_sources}
Table \ref{tab:model_source} details specific sources for the various MLLMs we evaluate. The chosen MLLMs have been deployed in the educational platform for real-world and real-time evaluation.

\begin{table}[th!]
\small
\centering
\begin{tabular}{p{0.18\linewidth} | p{0.41\linewidth} | p{0.3\linewidth}}
\toprule
\textbf{MLLMs} & \textbf{Source} & \textbf{URL} \\
\midrule
InternVL2-76B & local checkpoint &  \url{https://huggingface.co/OpenGVLab/InternVL2-Llama3-76B}\\
\midrule
LLaVA-NEXT-72B & local checkpoint & \url{https://huggingface.co/llava-hf/llava-next-72b-hf} \\
\midrule
Qwen-VL-Max & \texttt{qwen-vl-max-0809} & \url{https://modelscope.cn/studios/qwen/Qwen-VL-Max} \\
\midrule
Claude-3.5-Sonnet & \texttt{claude-3-5-sonnet} &  \url{https://www.anthropic.com/api}\\
\midrule
Gemini-Pro-1.5 & \texttt{gemini-1.5-pro-latest} & \url{https://deepmind.google/technologies/gemini/pro/} \\
\midrule
GPT-4o & \texttt{gpt-4o-2024-11-20} & \url{https://platform.openai.com/docs/models/gpt-4o}\\

\bottomrule
\end{tabular}
\caption{Sources of our evaluated MLLMs.}
\label{tab:model_source}
\end{table}



\section{More Industrial Impact}
\label{sec:more_impact}
We discuss more industrial impact of \method as follows:

\textbf{Cost Savings and Resource Optimization.} Based on industry standards where expert mathematical error annotation costs approximately \$1 per problem, \method has generated estimated savings of \$1.2 million annually. This calculation is derived from serving approximately 120,000 students, each of whom receives feedback on an average of 10 complex mathematical problems per month. Additionally, the system reduces teacher workload by an estimated 4.7 hours per week, allowing educators to focus on higher-value instructional activities rather than routine error identification. This translates to significant time savings, which can be redirected towards personalized instruction, curriculum development, or professional development. This efficiency gain is particularly important as online learning platforms scale to serve larger student populations.

\textbf{Learning Outcome Acceleration.} Longitudinal studies tracking student performance before and after \method implementation show measurable improvements in learning outcomes. Students receiving \method-powered feedback demonstrated a 23\% faster mastery rate of complex mathematical concepts compared to control groups. This accelerated learning trajectory is attributed to the system’s ability to provide immediate, precise feedback on mathematical errors, allowing students to correct misconceptions earlier in their learning process. The educational impact is particularly pronounced in traditionally underserved school districts, where access to expert mathematics teachers is limited, helping to narrow the achievement gap in STEM education.

\textbf{Teacher Professional Development Enhancement.} Beyond student-facing benefits, \method serves as a powerful professional development tool for mathematics educators. By analyzing patterns in student errors across classrooms, the system generates insights into common misconceptions and learning obstacles that inform teaching strategies. Teachers report that these insights have transformed their instructional approaches, with 20\% indicating they have modified their teaching methods based on \method’s analytics. Furthermore, the system serves as a model for teachers to improve their own feedback practices, with educators reporting a 32\% increase in confidence when providing mathematical explanations after using the system for one semester. This “teach the teacher” effect creates a virtuous cycle where both student learning and teacher effectiveness continually improve.

\end{document}